\documentclass{article}

\usepackage{microtype}
\usepackage{graphicx}
\usepackage{subcaption} 
\usepackage{booktabs}   

\usepackage{hyperref}

\usepackage{algorithm}
\usepackage{algorithmic} 

\usepackage[preprint]{icml2026}

\usepackage{amsmath}
\usepackage{amssymb}
\usepackage{mathtools}
\usepackage{amsthm}
\usepackage{xspace}
\usepackage{xcolor}
\usepackage{colortbl}
\usepackage{pifont}
\usepackage{multirow}
\usepackage{fancybox}
\usepackage{hhline}
\usepackage{tabularx}
\usepackage{bbm}
\usepackage{amsfonts}
\usepackage{pythonhighlight}
\usepackage[inline,shortlabels]{enumitem}
\usepackage{tikz}

\def\name{AutoScout\xspace}

\newcommand{\eg}{e.g.,\xspace}
\newcommand{\ie}{i.e.,\xspace}

\newenvironment{denseitemize}{
  \begin{itemize}[itemsep=0pt,parsep=0pt,topsep=0pt,partopsep=0pt,leftmargin=*]
}{\end{itemize}}
\graphicspath{{imgs/}{sections/imgs/}{sections/}}

\begin{document}

\twocolumn[
  \icmltitle{AutoScout: Structured Optimization for Automating ML System Configuration}

  \icmlsetsymbol{equal}{*}

  \begin{icmlauthorlist}
     \icmlauthor{Jimmy Shong}{aff1}
     \icmlauthor{Yuhan Ding}{aff1}
     \icmlauthor{Yihan Jiang}{aff1}
     \icmlauthor{Liheng Jing}{aff1}
     \icmlauthor{Haonan Chen}{aff3}
     \icmlauthor{Gaokai Zhang}{aff1}
     \icmlauthor{Aditya Akella}{aff2}
     \icmlauthor{Fan Lai}{aff1}
  \end{icmlauthorlist}

  

  \icmlaffiliation{aff1}{University of
Illinois Urbana-Champaign}

  \icmlaffiliation{aff3}{University of California, Berkeley}
  
  \icmlaffiliation{aff2}{The University of Texas at Austin}
  
  \icmlcorrespondingauthor{Fan Lai}{fanlai@illinois.edu}

  \icmlkeywords{Machine Learning, Cloud Computing}

  \vskip 0.3in
]

\printAffiliationsAndNotice{} 

\begin{abstract}
Machine learning (ML) systems expose a rapidly expanding configuration space spanning model-parallelism strategies, communication optimizations, and low-level runtime parameters. End-to-end system efficiency is highly sensitive to these choices, yet identifying high-performance configurations is challenging due to heterogeneous feature types (e.g., sparse and dense parameters), conditional dependencies (e.g., valid execution parameters only under specific upstream decisions), and the high search (profiling) cost. 
Existing approaches either optimize a narrow subset of configuration dimensions or rely on ad-hoc heuristics that fail to generalize as configuration spaces continue to grow. We present AutoScout, a general-purpose systems configurator for ML training, fine-tuning, and inference. It formulates the system configuration as a mixed-discrete/continuous optimization problem with hierarchical dependencies and introduces a hybrid optimization framework that jointly refines sparse structural decisions and dense execution parameters. To reduce profiling cost, AutoScout adaptively prioritizes high-impact configuration features and ensembles simulators with varying fidelity. Across diverse models, hardware platforms, and deployment objectives, AutoScout consistently identifies high-performance configurations, achieving 2.7--3.0$\times$ training speedup over expert-tuned settings.
\end{abstract}

\section{Introduction}
\label{sec:intro}

Machine learning systems expose an increasingly wide spectrum of performance-critical optimizations, spanning model parallelism~\citep{narayanan2019pipedream, zheng2022alpa}, communication overlap~\citep{gpipe}, and low-level runtime parameters (e.g., chunked-prefill size). Each new framework, hardware, or deployment objective introduces compound configuration dimensions whose interactions are already difficult to reason about in isolation. Worse yet, this configuration space expands rapidly with the prevalence of distributed deployments. Our analysis of widely used frameworks, such as vLLM~\citep{vllm} and Megatron~\citep{shoeybi2020megatronlmtrainingmultibillionparameter}, shows that even for a fixed model and hardware setup, the number of valid configurations that already reaches the order of thousands continues to grow year over year as frameworks evolve.

Our studies show that a well-configured deployment can improve end-to-end execution speed by 1.4--4.0$\times$ compared to naive or default settings. 
However, identifying such configurations is notoriously challenging. First, ML configuration spaces combine \emph{sparse} features (e.g., categorical choices such as parallelism strategies) with \emph{dense} features (e.g., continuous execution parameters such as DDP communication bucket size and GPU memory utilization). Second, these features are rarely independent: many parameters are only meaningful under specific upstream decisions (e.g., SM allocation for communication in tensor parallellism), leading to strong hierarchical and conditional dependencies. Third, the performance-optimal configuration is highly context-dependent; it varies across models, accelerator types and counts, and deployment objectives such as latency-throughput tradeoffs in serving~\citep{narayanan2019pipedream, piper}. Finally, evaluating a single configuration often requires expensive GPU-based execution, making efficient search essential.

Existing advances are fundamentally limited. First, they mostly focus on optimizing a narrow subset of configuration dimensions, such as data, tensor, or pipeline parallelism degrees~\citep{narayanan2019pipedream, shoeybi2020megatronlmtrainingmultibillionparameter, zheng2022alpa, metisatc24}. However, even under identical parallelism strategies, system performance can vary by up to $42\times$ depending on other execution parameters (\S\ref{sec:background}). Second, more recent advances rely on task-specific heuristics (\eg limiting tensor parallelism to single nodes or prioritizing data parallelism via capacity-aware pruning) \citep{piper, metisatc24, strati2025sailorautomatingdistributedtraining} or carefully engineered simulators, yet difficult to sustain as configuration spaces continue to evolve. Third, traditional black-box optimization methods, such as Bayesian Optimization, are ill-suited to such configuration spaces that combine sparse and dense parameters with strong hierarchical dependencies \citep{jamieson2016, li2018hyperbandnovelbanditbasedapproach}. Furthermore, they fail to account for the high-profile cost per search step.

In this paper, we introduce \name, a systems configuration optimizer for ML training, fine-tuning, and inference. Given a target model, deployment objective, and accelerator environment, \name efficiently navigates the configuration space to identify high-performance configurations. At its core, \name employs a hybrid optimization framework that jointly explores sparse structural decisions and refines dense execution parameters. Sparse configuration choices are organized into a tree-based search space explored by a sparse optimizer, while dense features are optimized using coordinate-wise stochastic gradient descent. Feedback from each search step is integrated through a hybrid bandit mechanism that dynamically coordinates exploration between the sparse and dense optimizers. To further improve efficiency, \name incorporates a tournament-based design to prioritize high-impact configuration features and adaptively ensembles multiple simulators with varying fidelity to reduce reliance on expensive profiling.

In summary, this paper makes the following contributions:
\begin{denseitemize}
    \item We identify ML systems configuration optimization as a mixed discrete-continuous optimization problem with strong hierarchical and conditional dependencies.

    \item We propose \name, a novel hybrid system configurator that combines tree-based search over sparse structural decisions with gradient-guided optimization of dense execution parameters, coordinated by adaptive exploration.

    \item We demonstrate that \name consistently outperforms existing approaches across diverse models and deployment objectives, generating configurations that lead to $1.3$--3.0$\times$ speedup while being 13.7--16.5$\times$ faster than existing system configurators.
\end{denseitemize}

\section{Background and Motivation}
\label{sec:background}

As model sizes continue to grow and training and inference pipelines increasingly exploit heterogeneous hardware, ML practitioners must navigate an ever-expanding configuration space spanning parallelism strategies, execution-level parameters, and system optimization policies~\cite{jitserve-nsdi26, hygen-neurips25}, e.g., using activation checkpointing or not~\citep{piper, zheng2022alpa}.

\begin{figure}[t]
    \centering
  \includegraphics[width=0.95\linewidth]{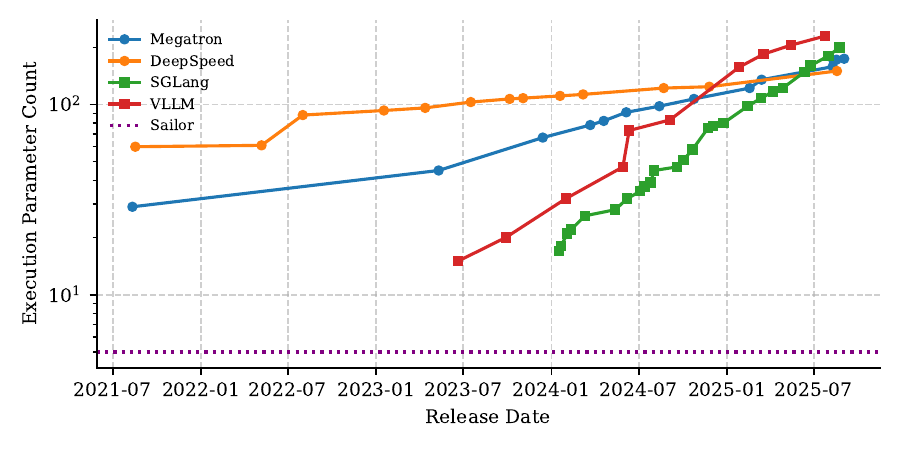}
    \caption{The number of exposed configuration knobs in modern ML systems has steadily increased over time, significantly expanding the dimensionality and complexity of the optimization space. }
    \label{fig:increasing_knobs}
\end{figure}

\paragraph{Ever-Growing Systems Configuration Complexity in ML Workloads.}
Even when fixing the underlying model and hardware, the space of valid system configurations can be enormous. A single deployment typically involves a mixture of discrete, sparse decisions (e.g., parallelism strategies), continuous parameters (e.g., GPU memory utilization and communication bucket size), and conditional constraints that tie these choices together. 
As shown in Figure~\ref{fig:increasing_knobs}, the number of exposed configuration parameters in widely used ML systems has increased steadily over time. Each new knob expands the search space and introduces additional dependencies that must be reasoned about jointly. 

\begin{figure}[t]
    \centering
  \includegraphics[width=0.8\linewidth]{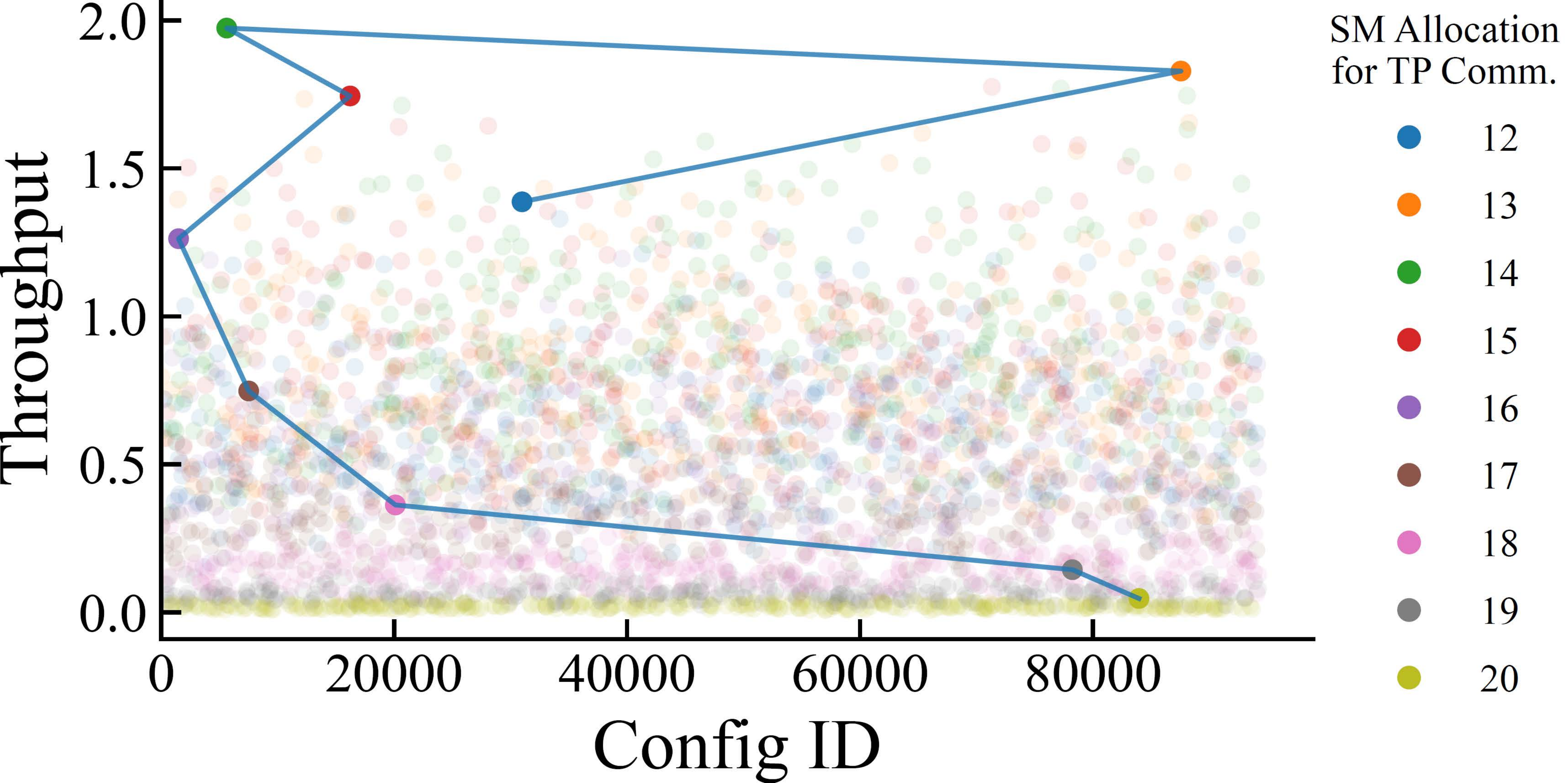}
    \caption{Configuration space in modern ML systems is enormous, consisting of both discrete and continuous features and leading to orders-of-magnitude throughput differences.}
    \label{fig:sensitive_knobs}
\end{figure}

Indeed, even on identical hardware, Figure~\ref{fig:sensitive_knobs} shows that the performance gap between two configurations can exceed $40\times$. Such disparities indicate that naive defaults or ad hoc tuning can leave substantial ML system efficiency untapped. At the scale of frontier-model deployment, even modest performance improvements can translate into jobs completing weeks or months earlier, yielding millions of dollars in cost savings~\citep{cottier2025risingcoststrainingfrontier}. 

Unfortunately, optimal system configurations are highly dependent on the underlying hardware, model characteristics, and deployment objectives, making it difficult for hand-crafted heuristics or static rules to generalize. For example, the optimal communication bucket size and parallelism strategies vary significantly across GPU types and network topologies~\citep{metisatc24}. 
Therefore, even when carefully engineered heuristics perform well initially, they quickly become outdated as frameworks, optimization stratgies, or the user's latency requirements evolve. 

\paragraph{Limitations of Existing Solutions.} 

Existing configuration optimizers for ML systems largely focus on a narrow subset of decisions, most notably three-dimensional model parallelism (data, tensor, and pipeline parallelism)~\cite{metisatc24,strati2025sailorautomatingdistributedtraining}. However, even when the best 3D parallelism strategy is selected for a fixed cluster, execution-level configuration choices can lead to performance gaps of 1.4-4$\times$ (\S\ref{eval:e2e}). These results indicate that parallelism-only optimization leaves substantial performance untapped.

To manage the resulting complexity, prior systems often rely on task-specific heuristics or carefully engineered simulators to prune the search space. Such approaches are inherently brittle: heuristics must be repeatedly redesigned as new execution optimizations, and simulator accuracy degrades rapidly as configuration interactions grow more complex (Figure~\ref{fig:increasing_knobs}). As a result, heuristic-driven search does not scale with the pace at which ML frameworks and hardware evolve. Yet, general black-box optimizers such as Bayesian optimization struggle due to high-dimensional, mixed discrete-continuous spaces, as well as prohibitive profiling cost~\cite{zheng2022alpa}. 

Together, these limitations expose a fundamental gap: existing approaches neither scale with configuration complexity nor adapt reliably across ML deployments.
\section{\name Design}
\label{sec:design}

This paper introduces \name, a general-purpose systems configurator  for ML training, fine-tuning, and inference.
\name aims to efficiently identify high-performance configurations in large, structured configuration spaces by explicitly reasoning about hierarchical dependencies, heterogeneous parameters, and high profiling cost. 

\subsection{Design Overview}
\label{subsec:overview}

Unlike existing approaches that rely on feature-specific heuristics or simulators~\citep{piper, metisatc24, strati2025sailorautomatingdistributedtraining}, \name decomposes configuration optimization into coordinated sparse structural exploration and dense continuous refinement. This separation enables scalable and extensible optimization as ML systems and configuration spaces continue to evolve.

\begin{figure}[t]
    \centering
  \includegraphics[width=0.99\linewidth]{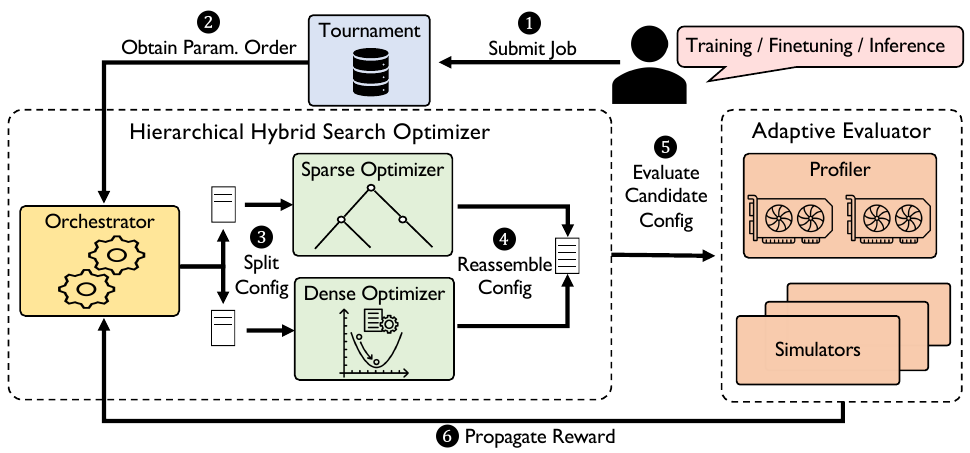}
    \caption{\name overview and workflow. It combines sparse and dense optimizers with an adaptive evaluator to efficiently search the ML system configuration space.}
    \label{fig:lifecycle}
\end{figure}

Figure~\ref{fig:lifecycle} illustrates the overall workflow of \name. Upon receiving a user request, the \emph{Orchestrator} coordinates the optimization process by dispatching sparse structural features and dense (continuous) features to the \emph{Sparse Optimizer} and \emph{Dense Optimizer}, respectively. Each optimizer proposes candidate configurations for the features under its control and forwards them to the \emph{Synthesizer}, which assembles complete configurations and determines whether to evaluate them using simulator ensembles or real profiling, based on confidence and estimated cost. Evaluation feedback is then returned to the Orchestrator to guide subsequent search steps, including refining feature-level importance. This iterative process continues until convergence criteria are met or a time budget is reached, at which point \name terminates the optimization loop and returns the selected configuration to the user.

We next introduce the design of the Sparse Optimizer and the Dense Optimizer to minimize the number of search steps (\S\ref{subsec:hybrid-optimizer}), while adaptively orchestrating simulators to minimize per-step search overhead (\S\ref{subsec:coordination}).

\subsection{Hierarchical Hybrid Search Optimizer}
\label{subsec:hybrid-optimizer}

Configuration optimization for ML systems differs fundamentally from classical hyperparameter tuning. Beyond the coexistence of sparse and dense features, many configuration decisions are hierarchical and conditional: downstream parameters are only meaningful once upstream structure is fixed~\citep{piper, zheng2022alpa}. For example, valid tensor- and data-parallelism degrees depend on how the model is partitioned into pipeline stages, while communication overlap depends on the activated parallelism policy. This induces a search space with strict ordering constraints and fragmented feasibility regions across sparse and dense dimensions, exceeding the capabilities of existing flat optimization strategies~\citep{ malouf_high_dim_bo}.

Our key insight is that sparse configuration features naturally form a dependency tree with categorical branching, making them amenable to tree-based exploration. In contrast, dense execution parameters often exhibit local smoothness within a fixed structure, allowing efficient numerical refinement. Motivated by this structural dichotomy, \name adopts a hierarchical hybrid optimization strategy that explicitly separates sparse structural exploration from dense parameter optimization, enabling efficient and scalable search in large, structured configuration spaces.

\paragraph{Sparse Optimizer with Adaptive Feature Prioritization.}
\name models sparse, structural configuration decisions as a dependency-aware search tree and explores this space using Monte Carlo Tree Search (MCTS)\citep{mcts}. 
Each node corresponds to a partial configuration $\mathbf{s} \in \mathcal{S}$, and each edge represents a valid refinement that satisfies feasibility constraints induced by upstream decisions.
MCTS naturally supports conditional validity, delayed rewards, and asymmetric branching factors, making it well-suited for hierarchical configuration spaces.
Rollouts complete partial configurations by invoking downstream numerical optimization and return an observed reward $r(\mathbf{s})$, such as vLLM's per-token latency with the proposed configuration, for backpropagation.

However, not all configuration features contribute equally to performance.
Let $\mathcal{F}=\{f_1,\ldots,f_d\}$ denote the set of sparse structural features.
In practice, only a small subset $\mathcal{F}^\star \subset \mathcal{F}$ has significant performance impact for a given workload~\citep{selecta, 201567}, and the identity of $\mathcal{F}^\star$ varies across models, hardware environments, and deployment objectives, so pre-engineering a fixed feature ordering is impractical.
Moreover, the ordering of features in the search tree critically affects MCTS efficiency: prioritizing high-impact features earlier reduces effective branching and accelerates credit assignment, while suboptimal orderings lead to wasted exploration.

To address this challenge, \name introduces a lightweight \emph{tournament-based feature prioritization} mechanism that adapts the tree structure online.
Our key observation is that while feature importance is job-dependent, it often lies within a small set of commonly effective orderings.
Accordingly, \name maintains a candidate set of $K$ tree structures $\{\mathcal{T}_1,\ldots,\mathcal{T}_K\}$, each corresponding to a different prioritization of $\mathcal{F}$ and initialized from prior optimization runs (e.g., past jobs).
During a brief warm-up phase, these trees are instantiated with negligible overhead.

At each optimization iteration, \name alternates among the candidate trees, allowing each $\mathcal{T}_k$ to propose a configuration $\mathbf{s}_k$.
The selected configuration is evaluated once, and the resulting reward is propagated to all $K$ trees, ensuring shared learning without duplicated evaluation cost.
To avoid bias from update order, \name traverses trees in a zigzag schedule across rounds, ensuring balanced credit assignment. 
After each tournament round, which consists of $K$ optimization iterations, \name ranks tree structures by the cumulative reward of their proposed configurations and retains only the top-performing half.
This elimination process repeats until a single tree structure remains.
The tournament converges in $O(\log K \cdot N)$ iterations and incurs minimal overhead, as each iteration requires only one configuration evaluation.
Our evaluations show that this mechanism rapidly identifies effective feature orderings, significantly improving MCTS efficiency (\S\ref{eval:ablation}).

\paragraph{Dependency-aware Dense Optimizer.}
Given a fixed structural configuration, system performance still depends critically on continuous execution parameters, such as chunked-prefill length, kernel fusion thresholds, and GPU memory budget for KV cache.
These parameters often exhibit local smoothness, but their feasible ranges and performance impact can depend on other decisions (e.g., KV cache offloading or not determined by the sparse optimizer), making global surrogate modeling unreliable.

Structural decisions may activate, constrain, or disable subsets of dense parameters. Rather than rejecting proposals from either optimizer, \name explicitly encodes these dependencies using a structure-dependent masking function $M(\mathbf{s})$, which maps the global dense parameter space to a feasible subspace $\mathcal{X}(\mathbf{s})$. When the sparse optimizer proposes a new structure $\mathbf{s}$, the dense optimizer projects its current state onto $\mathcal{X}(\mathbf{s})$ by masking inactive dimensions, naturally aligning with its coordinate-wise update strategy.

The dense optimizer employs a coordinate-wise search strategy that refines one parameter at a time using a momentum-based update rule. At each iteration, the optimizer perturbs the active parameter along its current search direction; if the perturbation yields improvement, it continues along that direction to exploit local structure, while failures trigger a switch to a different parameter to encourage exploration across dimensions. This design naturally accommodates the discrete, grid-structured nature of execution parameters exposed by modern ML systems, where valid settings are typically drawn from a finite set of system-supported values rather than a continuous range.

However, end-to-end performance is jointly determined by structural and dense parameters, making the reward $r(\mathbf{s}, \mathbf{x})$ non-separable.
Poor performance may stem from poorly tuned dense parameters, and naïvely attributing the reward to the sparse optimizer can mislead learning.
To address this challenge, we next introduce an \emph{optimizer orchestrator} that mediates reward assignment and controls optimization flow.

\subsection{Optimizer Orchestrator}
\label{subsec:coordination}

Our optimizer orchestrator adaptively decomposes the reward signal $r(s)$ to inform the update of the sparse and dense optimizers, based on observed improvement trends and uncertainty estimates. In addition, it dynamically selects between low-cost simulator-based evaluation and high-fidelity profiling for each proposed configuration, exposing a fundamental tradeoff between feedback fidelity and per-step efficiency: over-reliance on noisy simulator feedback can misguide optimizer updates and increase the number of search iterations, while excessive reliance on profiling incurs prohibitively high evaluation cost.

\begin{algorithm}[t]
\caption{\name Configuration Optimization Workflow}
\label{alg:workflow}
\begin{algorithmic}[1]
\STATE \textbf{Input:} workload $\mathcal{W}$, search budget $T$
\STATE \textbf{Hyperparams:} check interval $\tau$, error threshold $\epsilon$
\STATE \textbf{Output:} best configuration $(S^\star, X^\star)$

\STATE \textcolor{blue}{\footnotesize \itshape // Phase 0: Tournament warm-start (optional)}
\STATE $(\mathcal{T}^\star, \pi^\star) \gets \textsc{TournamentWarmStart}(\mathcal{W})$\label{algo:tournament}

\STATE \textcolor{blue}{\footnotesize \itshape // Phase 1: Initialization}
\STATE $\mathcal{O} \gets \textsc{InitOrchestrator}(\tau,\epsilon)$
\STATE $\mathcal{S} \gets \textsc{InitSparseOptimizer}(\mathcal{T}^\star,\pi^\star)$
\STATE $\mathcal{D} \gets \textsc{InitDenseOptimizer}(\mathcal{W})$
\STATE $(S_{\text{base}}, S_{\text{cand}}) \gets \textsc{ProposeSparsePair}(\mathcal{S})$
\STATE $(X_{\text{base}}, X_{\text{cand}}) \gets \textsc{ProposeDensePair}(\mathcal{D})$
\STATE $(S^\star, X^\star) \gets (S_{\text{base}}, X_{\text{base}})$

\STATE \textcolor{blue}{\footnotesize \itshape // Phase 2: Iterative coordinated optimization}
\FOR{$t \gets 1$ to $T$}
    \STATE \textcolor{gray}{\footnotesize \itshape // Construct a $2\times2$ evaluation batch}
    \STATE $C_{bb}, C_{bc}  \gets (S_{\text{base}}, X_{\text{base}}), (S_{\text{base}}, X_{\text{cand}})$\label{algo:batch_begin}
    \STATE $C_{cb}, C_{cc}  \gets (S_{\text{cand}}, X_{\text{base}}), (S_{\text{cand}}, X_{\text{cand}})$
    \label{algo:batch_end}
    \STATE $\mathcal{C} \gets \{C_{bb}, C_{bc}, C_{cb}, C_{cc}\}$

    \STATE \textcolor{gray}{\footnotesize \itshape // Fidelity-adaptive evaluation (simulators vs.\ profiling)}
    \STATE $\mathbf{c} \gets \textsc{EvalBatchAdaptive}(\mathcal{O}, \mathcal{C}, t)$\label{algo:eval_adaptive}

    \STATE \textcolor{gray}{\footnotesize \itshape // Update best-so-far and orchestrator state}
    \STATE $(S^\star, X^\star) \gets \textsc{UpdateBest}((S^\star,X^\star), \mathcal{C}, \mathbf{c})$
    \STATE $\mathcal{O} \gets \textsc{UpdateOrchestrator}(\mathcal{O}, \mathcal{C}, \mathbf{c})$\label{algo:update_orchestrator}

    \STATE \textcolor{gray}{\footnotesize \itshape // Orchestrated candidate generation for next round}
    \STATE $(S_{\text{base}}, S_{\text{cand}}) \gets \textsc{NextSparsePair}(\mathcal{O}, \mathcal{S})$\label{algo:next_sparse}
    \STATE $(X_{\text{base}}, X_{\text{cand}}) \gets \textsc{NextDensePair}(\mathcal{O}, \mathcal{D})$\label{algo:next_dense}
\ENDFOR
\STATE \textbf{return} $(S^\star, X^\star)$
\end{algorithmic}
\end{algorithm}

Algorithm~\ref{alg:workflow} illustrates \name's configuration optimization workflow.
We optionally warm-start the sparse search via a tournament over candidate tree structures (\ie parameter ordering), retaining a high-performing structure (Line~\ref{algo:tournament}).
The optimization then proceeds iteratively under the coordination of an optimizer orchestrator.
At each iteration, \name forms a small evaluation batch by combining the current baseline and candidate configurations proposed by the sparse and dense optimizers (Lines~\ref{algo:batch_begin}--\ref{algo:batch_end}). 
The batch is evaluated through a fidelity-adaptive controller that dynamically selects between low-cost simulator feedback and high-fidelity profiling (Line~\ref{algo:eval_adaptive}).
Based on the evaluation results, the orchestrator updates its internal state and coordinates the generation of the next sparse and dense candidate configurations for the subsequent iteration (Lines~\ref{algo:update_orchestrator}--\ref{algo:next_dense}).

\paragraph{Coordinating Sparse and Dense Optimizers.}

\name coordinates the sparse and dense optimizers using a hierarchical multi-armed bandit formulation~\citep{jamieson2016}. 
Each optimizer is modeled as an arm, where pulling an arm corresponds to allocating one unit of search budget, i.e., an optimization iteration that proposes and evaluates a candidate configuration. 
A bandit formulation is well suited for this setting because the relative benefit of exploring new structural configurations versus refining dense execution parameters is unknown \emph{a priori}, varies across workloads, and evolves over the course of optimization.

At iteration $t$, the orchestrator selects an optimizer $a_t \in \{\textsc{Sparse}, \textsc{Dense}\}$ according to the Upper Confidence Bound (UCB1) criterion:
\begin{equation}
\label{eq:ucb}
a_t = \arg\max_{a} \left[
    \frac{Q_a}{N_a}
    + C(t)\sqrt{\frac{\ln N_{\text{total}}}{N_a}}
\right],
\end{equation}
where $Q_a$ and $N_a$ denote the cumulative reward and number of selections for optimizer $a$, respectively, and $N_{\text{total}}$ is the total number of iterations so far. 
The exploration coefficient $C(t)=C_0 \cdot \gamma^{t}$ decays exponentially over time, enabling the orchestrator to favor exploration early in the search and gradually shift toward exploitation as confidence in promising optimization directions increases.

The bandit observes reward signals derived from a difference-of-differences estimator that disentangles the marginal contributions of each optimizer.
Specifically, at each iteration, the orchestrator evaluates a $2 \times 2$ grid of configurations combining sparse baseline/candidate with dense baseline/candidate, computing marginal improvements $\Delta_{\textsc{Sparse}}$ and $\Delta_{\textsc{Dense}}$ while controlling for the other optimizer's contribution.
Early in the search, the bandit favors structural, sparse features to rapidly prune unpromising regions of the configuration space; as promising structures emerge, it shifts budget toward dense refinement to exploit local smoothness.
This adaptive exploration--exploitation tradeoff enables \name to dynamically balance global structural search and local parameter optimization without manual scheduling, improving robustness and sample efficiency across workloads with heterogeneous structure-parameter interactions.

\paragraph{Cost-Aware Profiling with Simulators.}
Evaluating a configuration via real profiling is time- and resource-intensive, often requiring dedicated GPU execution. To reduce per-iteration overhead, \name maintains an ensemble of lightweight simulators that provide fast performance predictions based on subsets of configuration features. During optimization, simulator predictions are aggregated to estimate configuration cost, enabling efficient early-stage search and reducing reliance on expensive real measurements.

To guard against simulator bias, \name periodically validates selected configurations using real profiling at fixed intervals of $\tau$ iterations, computing the mean absolute percentage error (MAPE) between predicted and observed costs. When the error exceeds a threshold $\epsilon$, \name triggers a fidelity switch that transitions subsequent evaluations from simulation to real profiling.

This transition preserves accumulated search knowledge rather than discarding it. The MCTS tree structure is retained. Bandit statistics collected during simulation are treated as weak priors rather than reset, allowing the optimizer to adapt smoothly to higher-fidelity feedback. In addition, the top-$K$ configurations identified during simulation are prioritized for re-evaluation under real profiling. Together, these mechanisms ensure that simulators provide a meaningful head start by narrowing the search space, while the adaptive fidelity switch prevents inaccurate predictions from misleading subsequent optimization.
\section{Evaluation}
\label{sec:eval}

We evaluate \name using a cluster of 8 A100s and 4 A40 GPUs, spanning LLM training and inference jobs. Our evaluations show that \name generates configurations that leads to $1.3$--3.0$\times$ speedup while using $28.6\%$--$93.07\%$ less search steps than existing advances.

\subsection{Evaluation Setup}

\paragraph{Models and Workloads.}
We evaluate \name across models with diverse architectures and scales, including \textsc{Llama-3.2-3B}, \textsc{Llama-3.1-Nemotron-Nano-VL-8B-V1}, and the \textsc{Qwen3-30B-A3B} variant as training jobs on the LMSYS-Chat-1M \cite{zheng2023lmsyschat1m} dataset. For inference, we evaluate on \textsc{Meta-Llama-3-8B-Instruct} using the LMSYS-Chat-1M dataset.
For each model, \name searches a large, multi-dimensional configuration space spanning tensor-parallel degree, micro-batch size, communication bucket size, activation recomputation, and other widely used execution parameters, resulting in up to $\sim$30{,}000 feasible configurations. 
Following prior work~\citep{strati2025sailorautomatingdistributedtraining, metisatc24}, we construct lightweight model latency simulators based on multiple linear regression to enable efficient performance estimation during search. Additional details on the configuration space and simulator design are provided in Appendix \ref{appendix:megatron_knobs} and \ref{appendix:simulators}, respectively.

\paragraph{Baselines.}
We compare against the following advances:
\begin{denseitemize}
    \item \emph{vLLM}~\cite{vllm}: a widely used LLM serving framework with expert-tuned default configurations.

    \item \emph{Megatron-LM}~\cite{shoeybi2020megatronlmtrainingmultibillionparameter}: a popular distributed LLM training framework using expert-recommended parallelism and execution settings.

    \item \emph{UDO}~\cite{wang2021udouniversaldatabaseoptimization}: a general-purpose system configuration optimizer that applies MCTS to explore configuration spaces.

    \item \emph{CherryPick}~\cite{201567}: a cloud job configuration framework based on Bayesian Optimization.

    \item \emph{Metis}~\cite{metisatc24}: a system that automatically discover the best 3D model parallelism plan for machine learning jobs on heterogeneous clusters.

\end{denseitemize}

\paragraph{Metrics.} 
Our evaluation focuses on two objectives: (i) minimizing configuration search time and profiling cost, while (ii) maximizing end-to-end system performance.
For training workloads, we report execution latency measured in seconds per training iteration ($\mathrm{s/iter}$).
For inference workloads, we report generation latency measured in milliseconds per generated token ($\mathrm{ms/token}$).

All results are reported as averages over 20 independent runs to ensure statistical significance.

\subsection{End-to-End Performance}
\label{eval:e2e}

\begin{figure}[t]
  \centering
  \begin{minipage}{0.235\textwidth}
    \centering
    \includegraphics[width=\linewidth]{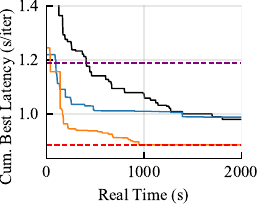}
  \end{minipage}
  \hfill
  \begin{minipage}{0.235\textwidth}
    \centering
    \includegraphics[width=\linewidth]{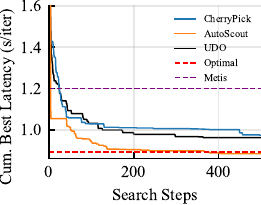}
  \end{minipage}
  \caption{
  End-to-end training performance and search behavior of \name on \textsc{Qwen-MoE}.
  }
  \label{fig:e2e_qwen}
\end{figure}

\begin{figure}[t]
  \centering
  \begin{minipage}{0.235\textwidth}
    \centering
    \includegraphics[width=\linewidth]{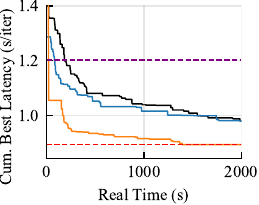}
  \end{minipage}
  \hfill
  \begin{minipage}{0.235\textwidth}
    \centering
    \includegraphics[width=\linewidth]{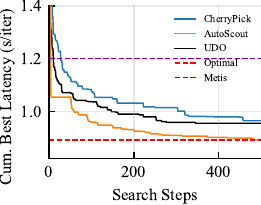}
  \end{minipage}
  \caption{
  End-to-end training performance and search behavior of \name on \textsc{Llama-3.2-3B}.
  }
  \label{fig:e2e_llama}
\end{figure}

\begin{figure}[t]
  \centering
  \begin{minipage}{0.235\textwidth}
    \centering
    \includegraphics[width=\linewidth]{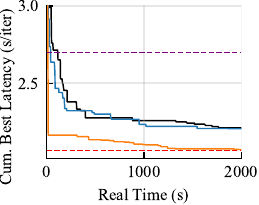}
  \end{minipage}
  \hfill
  \begin{minipage}{0.235\textwidth}
    \centering
    \includegraphics[width=\linewidth]{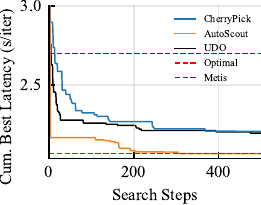}
  \end{minipage}
  \caption{
  End-to-end training performance and search behavior of \name on \textsc{Llama-3.1-Nemotron-Nano-VL-8B-V1}.
  }
  \label{fig:e2e_llamavl}
\end{figure}

\begin{figure}[t]
  \centering

  \begin{minipage}{0.235\textwidth}
    \centering
    \includegraphics[width=\linewidth]{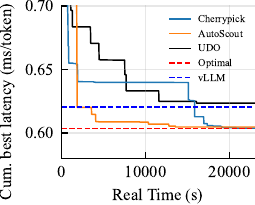}
    \caption{End-to-end inference performance of \name on \textsc{Qwen-MoE}}
    \label{fig:e2e_infer}
  \end{minipage}
  \hfill
  \begin{minipage}{0.235\textwidth}
    \centering
    \includegraphics[width=\linewidth]{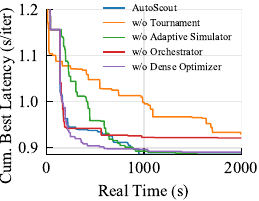}
    \caption{Performance breakdown of \name.}
    \label{fig:breakdown}
  \end{minipage}
\end{figure}

\paragraph{\name identifies high-performance configurations for ML systems.}
Figures~\ref{fig:e2e_qwen}--\ref{fig:e2e_infer},show that \name consistently identifies configurations with lower end-to-end latency than expert-tuned system configurations and prior automatic parallelization systems across all evaluated models. 
For the \textsc{Qwen-MoE} workload, \name's best configuration improves performance by 1.4--2.7$\times$. 
Similarly, when training \textsc{Llama-3.2-3B}, \name achieves a 1.3--2.7$\times$ improvement over Megatron. 
For \textsc{Llama-3.1-Nemotron-Nano-VL-8B-V1}, \name identifies a configuration that outperforms those generated by 3D auto-parallelizers 1.3--3.0$\times$, respectively. For inference jobs on \textsc{Qwen-MoE}, \name identifies a configuration that outperforms expert-tuned system defaults of vLLM by 1.02$\times$.

These gains arise because \name optimizes a broader set of performance-critical configuration dimensions beyond traditional three-dimensional model parallelism~\citep{narayanan2019pipedream, shoeybi2020megatronlmtrainingmultibillionparameter}, many of which have a substantial impact on execution efficiency but are not captured by existing auto-parallelization systems.

\paragraph{\name reliably converges to optimal configurations.}
Figures~\ref{fig:e2e_qwen}--\ref{fig:e2e_infer} further demonstrate \name's ability to consistently converge to the best-performing configurations for each job. In contrast, CherryPick and UDO often stagnate early in the search and plateau at suboptimal regions of the configuration space.  
For the \textsc{Qwen-MoE} workload, \name's best configuration outperforms CherryPick's best result by $1.10\times$ and UDO's by $1.06\times$. Similarly, when training \textsc{Llama-3.2-3B}, \name achieves a $1.08\times$ improvement. 
For \textsc{Llama-3.1-Nemotron}, \name identifies a configuration that exceeds those produced by CherryPick and UDO by $1.07\times$ and $1.06\times$, respectively. 

\paragraph{\name achieves superior search efficiency.}
Beyond identifying better final configurations, Figures~\ref{fig:e2e_qwen}--\ref{fig:e2e_infer} show that \name substantially improves search efficiency in both the number of optimization steps and wall-clock time. For the \textsc{Qwen-MoE} workload, \name reduces the number of search steps required to reach the best configuration by $87.1\%$ relative to CherryPick and by $91.7\%$ relative to UDO. When training \textsc{Llama-3.2-3B}, \name achieves step reductions of $80.0\%$ and $80.8\%$ compared to CherryPick and UDO, respectively. For \textsc{Llama-3.1-Nemotron-Nano-VL-8B-V1}, \name reduces the number of search steps by over $92\%$ against cloud config baselines. For inference jobs on \textsc{Qwen-MoE}, \name reduces the number of search steps by over $28.6\%$.

These reductions translate directly into substantial wall-clock speedups. \name achieves up to $16.5\times$ faster search time on \textsc{Qwen-MoE}, $13.7\times$ on \textsc{Llama-3.2-3B}, and $22.9\times$ on \textsc{Llama-3.1-Nemotron-Nano-VL-8B-V1}. Moreover, \name consistently outperforms competing methods throughout most of the search trajectory, exhibiting strong anytime performance and remaining effective under both limited and extended search budgets, an important property for online and production settings such as cloud deployments or large-scale cluster scheduling~\citep{oppertune}. 

\subsection{Ablation Studies}
\label{eval:ablation}

\paragraph{Performance Breakdown.} 
Figure~\ref{fig:breakdown} presents an ablation study of \name on \textsc{Qwen-MoE} training, where individual system components are selectively disabled under a fixed optimization budget of 2000 seconds. When \name is restricted to the sparse optimizer alone, it makes reasonable initial progress but settles on a configuration that is $1.46\times$ slower. This behavior reflects convergence to a suboptimal region and stems from the sparse optimizer's limited ability to exploit local smoothness in continuous, performance-sensitive parameters.

Reintroducing the dense optimizer while removing the orchestrator further degrades both search efficiency and final configuration quality, underscoring the orchestrator's critical role in adaptively coordinating sparse exploration and dense refinement based on their effectiveness at different stages of the search. Finally, when \name's evaluator is restricted to real profiling without simulator assistance, the system still converges but does so $1.19\times$ more slowly, reflecting the loss of low-cost feedback during early exploration.

\begin{figure*}[t]
\centering
\begin{minipage}{0.3\textwidth}
    \centering
    \includegraphics[width=.8\linewidth]{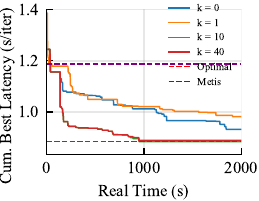}
    \caption{Moderate tournament sizes achieve the best trade-off between search efficiency and final convergence quality.}
    \label{fig:ablation-tournament}
\end{minipage}
\hfill
\begin{minipage}{0.3\textwidth}
    \centering
    \includegraphics[width=.85\linewidth]{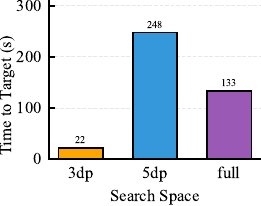}
    \caption{\name achieves fast planning for different configuration spaces.}
    \label{fig:ablation-space}
\end{minipage}
\hfill
\begin{minipage}{0.3\textwidth}
    \centering
    \includegraphics[width=.85\linewidth]{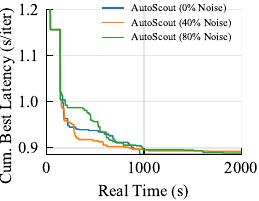}
    \caption{\name is robust to simulators of different accuracies.}
    \label{fig:ablation-simulator}
\end{minipage}
\end{figure*}

\paragraph{Impact of Number of Tournament Candidates.}
Figure~\ref{fig:ablation-tournament} presents an ablation study on the number of initial candidate tree structures $K$ used in \name's tournament-based feature prioritization. The results reveal a clear trade-off between early-stage exploration efficiency and tournament overhead. Small tournaments ($K{=}5$) already deliver substantial improvements over single- or no-tournament baselines, rapidly converging to near-optimal configurations by identifying high-impact feature orderings early in the search. In contrast, overly large tournaments ($K{=}40$) exhibit slower early progress due to the dilution of exploration budget across many candidate trees, although they eventually converge to competitive solutions. 
Overall, this ablation reports that \name can efficiently learn effective feature orderings with only a few candidates.

\paragraph{Scalability Across Configuration Spaces.}
Figure~\ref{fig:ablation-space} evaluates \name's scalability as the configuration space dimensionality increases. We consider three progressively larger feature spaces: a low-dimensional space (\texttt{3dp}) comprising knobs used in three-dimensional automatic parallelism; an intermediate space (\texttt{5dp}) that additionally includes five-dimensional parallelism degrees; and the full space (\texttt{full}), which further incorporates execution-level parameters such as \{\textit{ar}, \textit{ddp\_optim}, \textit{ddp\_bucket}\}.

Figure~\ref{fig:ablation-space} shows that despite the increase in search dimensionality, \name incurs only a modest increase in convergence time when scaling to larger configuration spaces. In some cases, \name converges even faster, as the richer feature space provides additional performance signals that enable more effective pruning and refinement.

\paragraph{Impact of Simulator Fidelity.}
Figure~\ref{fig:ablation-simulator} reports the robustness of \name's adaptive evaluator under increasing levels of noise injected into simulator outputs. We evaluate \name with 0\%, 40\%, and 80\% additive noise applied to simulator predictions, emulating progressively less reliable performance models. As simulator noise increases, we observe a gradual degradation in search efficiency, reflected in slower convergence toward low-latency configurations. 
Nevertheless, even under severe noise (80\%), \name remains stable and converges to near-optimal configurations by adaptively falling back to profiling-based evaluation. This behavior demonstrates the effectiveness of \name's adaptive evaluator in mitigating erroneous predictions and preventing catastrophic search failures. 
\section{Related Work}
\label{sec:related}

\paragraph{Cloud Configuration.}
Prior work on cloud configuration has largely focused on traditional analytics rather than modern ML systems. 
\textsc{Selecta}~\citep{selecta} formulates cloud storage configuration as a recommendation problem using collaborative filtering, while \textsc{CherryPick}~\citep{201567} applies Bayesian Optimization (BO) to select VM instance types.

While effective in their target domains, these approaches assume relatively flat configuration spaces and inexpensive evaluation and do not account for hierarchical dependencies, mixed sparse-dense features, or high profiling costs in ML system configuration~\citep{metisatc24, strati2025sailorautomatingdistributedtraining}.

\paragraph{Model Parallelism.}
Distributed training frameworks such as Megatron-LM and DeepSpeed~\citep{shoeybi2020megatronlmtrainingmultibillionparameter, rasley2020deepspeed} have demonstrated the effectiveness of large-scale parallel training, but require manually selecting parallelism degrees, often leading to suboptimal performance.
Automatic parallelization systems~\citep{zheng2022alpa, li2022amp, miao2022galvatron, piper, circinus-arxiv25} aim to automate this process and can identify near-optimal parallelism strategies under fixed assumptions.
To reduce search overhead, these systems typically rely on heuristics, simulators, or restricted search spaces.
However, most are designed for homogeneous clusters and pre-training workloads, and struggle to generalize to broader configuration dimensions~\citep{metisatc24}.

\paragraph{Tuning Algorithms and Search Methods.}
Monte Carlo Tree Search (MCTS)~\citep{mcts} has been applied to design-space exploration in ML systems.
\textsc{AlphaX}~\cite{wang2019alphax} employs MCTS to discover high-accuracy neural architectures with significantly fewer evaluations than prior NAS methods, while \textsc{LA-MCTS}~\cite{wang2020lamcts} accelerates optimization in high-dimensional black-box spaces by partitioning the search space.

\name builds on these insights by integrating MCTS with dense optimization, adaptive orchestration, and cost-aware evaluation to address the unique challenges of ML systems.
\section{Conclusion}
\label{sec:outro}
This paper presents \name, a general-purpose system configurator that addresses the growing complexity of ML system configuration across training, fine-tuning, and inference. By formulating configuration optimization as a structured mixed discrete-continuous problem, \name combines a hierarchical sparse optimizer with an efficient dense optimizer, coordinated through adaptive orchestration and cost-aware evaluation. Our evaluations show that \name identifies configurations with 1.3--3$\times$ speedup while being 16.5$\times$ faster than existing system configurators.

\bibliographystyle{icml2026}
\bibliography{main} 

\newpage
\appendix
\onecolumn
\section{Tuned Configuration Parameters}
\label{appendix:megatron_knobs}

Table~\ref{tab:megatron_knobs} lists the Megatron-LM configuration parameters that \name tunes during the optimization process for our training projects.

\begin{table}[ht]
\centering
\caption{Megatron-LM Parameters.}
\label{tab:megatron_knobs}
\small
\begin{tabular}{lll}
\toprule
\textbf{Configuration Parameter} & \textbf{Type} & \textbf{Range} \\
\midrule
\texttt{pp} & Int & $\{1, 2, 4, 8\}$ \\
\texttt{tp} & Int & $\{1, 2, 4, 8\}$ \\
\texttt{dp} & Int & $\{1, 2, 4, 8\}$ \\
\texttt{ep} & Int & $\{1, 2, 4, 8\}$ \\
\texttt{cp} & Int & $\{1, 2, 4, 8\}$ \\
\texttt{sp} & Bool & $\{\text{True}, \text{False}\}$ \\
\texttt{ar} & Bool & $\{\text{True}, \text{False}\}$ \\
\texttt{mbs} & Int & $\{1, 2, 4, 8\}$ \\
\texttt{ddp} & Int & $[1, \text{\#GPUs}]$ \\
\texttt{tp\_comm} & Int & $[12, 20]$ \\
\texttt{ddp\_bucket} & Int & $[1, 8]$ \\
\bottomrule
\end{tabular}
\end{table}

\noindent\textbf{Parameter Descriptions:}
\begin{itemize}[leftmargin=*,itemsep=2pt,parsep=0pt]
    \item \texttt{pp}: Pipeline parallelism degree, partitioning model layers across GPU groups.
    \item \texttt{tp}: Tensor parallelism degree, splitting individual layers across GPUs.
    \item \texttt{dp}: Data parallelism degree, replicating the model across GPU groups.
    \item \texttt{ep}: Expert parallelism degree for Mixture-of-Experts models.
    \item \texttt{cp}: Context parallelism degree for distributing long sequences.
    \item \texttt{sp}: Sequence parallelism, enabling parallel computation along the sequence dimension (requires \texttt{tp} $> 1$).
    \item \texttt{ar}: Activation recomputation/checkpointing to trade compute for memory.
    \item \texttt{mbs}: Micro batch size for gradient accumulation.
    \item \texttt{ddp}: Distributed data parallel bucket count (requires \texttt{dp} $> 1$).
    \item \texttt{tp\_comm}: The number of SMs allocated to communication in tensor parallelism.
    \item \texttt{ddp\_bucket}: the data-parallel communication bucket size in MB.
\end{itemize}

\section{Simulator Configurations}
\label{appendix:simulators}

Table~\ref{tab:simulators} lists the simulators used in our adaptive simulators experiment. Each simulator is a linear regression model trained on a different subset of configuration knobs, enabling the ensemble to capture diverse aspects of system performance.

\begin{table}[ht]
\centering
\caption{Simulators Used in the Adaptive Ensemble.}
\label{tab:simulators}
\small
\begin{tabular}{ll}
\toprule
\textbf{Simulator} & \textbf{Input Knobs} \\
\midrule
\textsf{3D-Parallelism} & \texttt{a100}, \texttt{a40}, \texttt{mbs}, \texttt{tp}, \texttt{pp}, \texttt{dp} \\
\textsf{5D-Parallelism} & \texttt{a100}, \texttt{a40}, \texttt{mbs}, \texttt{tp}, \texttt{pp}, \texttt{dp}, \texttt{ep}, \texttt{cp}, \texttt{sp} \\
\textsf{DDP-Aware} & \texttt{a100}, \texttt{a40}, \texttt{mbs}, \texttt{tp}, \texttt{pp}, \texttt{dp}, \texttt{ddp\_optim} \\
\textsf{Communication-Aware} & \texttt{a100}, \texttt{a40}, \texttt{mbs}, \texttt{tp}, \texttt{pp}, \texttt{dp}, \texttt{ar}, \texttt{tp\_comm} \\
\bottomrule
\end{tabular}
\end{table}

\noindent\textbf{Simulator Descriptions:}
\begin{itemize}[leftmargin=*,itemsep=2pt,parsep=0pt]
    \item \textsf{3D-Parallelism}: Models the baseline parallelism configuration with hardware allocation, micro-batch size, and core parallelism dimensions (tensor, pipeline, data).
    \item \textsf{5D-Parallelism}: Extends the core model with additional parallelism strategies including expert parallelism (\texttt{ep}), context parallelism (\texttt{cp}), and sequence parallelism (\texttt{sp}).
    \item \textsf{DDP-Aware}: Incorporates distributed data parallel optimization settings to model gradient synchronization overhead.
    \item \textsf{Communication-Aware}: Captures communication overhead through activation recomputation (\texttt{ar}) and tensor parallel communication overlap (\texttt{tp\_comm}).
\end{itemize}

\noindent During search, the ensemble prediction is computed as a weighted average of individual simulator outputs, where weights are determined by each simulator's $R^2$ score on held-out validation data:
\begin{equation}
    \hat{c}_{\text{ensemble}} = \sum_{i=1}^{4} w_i \cdot \hat{c}_i, \quad \text{where} \quad w_i = \frac{\max(0, R^2_i)}{\sum_{j=1}^{4} \max(0, R^2_j)}
\end{equation}

\end{document}